\documentclass[10pt,twocolumn,letterpaper]{article}

\usepackage{wacv}
\usepackage{amsmath,graphicx,epsfig,textcomp}

\usepackage{epstopdf}
\DeclareGraphicsExtensions{.eps}

\usepackage{amsmath,amssymb,multirow}
\usepackage{algorithm,algorithmic}
\usepackage[noadjust,nospace]{cite}

\usepackage{booktabs,enumerate}
\usepackage{url}
\usepackage{tabularx}
\usepackage{times}

 \normalsize

\wacvfinalcopy 

\def\wacvPaperID{226} 

\ifwacvfinal\fi
\begin{document}

\title{On the Importance of Normalisation Layers in Deep Learning with Piecewise Linear Activation Units}

\author{Zhibin Liao and Gustavo Carneiro \\
ARC Centre of Excellence for Robotic Vision, University of Adelaide, Australia \thanks{This research was supported by the Australian Research Council Centre of Excellence for Robotic Vision (project number CE140100016)} \\
{\tt\small \{zhibin.liao,gustavo.carneiro\}@adelaide.edu.au} }

%


\maketitle
\ifwacvfinal\thispagestyle{empty}\fi

\begin{abstract}

Deep feedforward neural networks with piecewise linear activations are currently producing the state-of-the-art results in several public datasets (e.g., CIFAR-10, CIFAR-100, MNIST, and SVHN).  
The combination of deep learning models and piecewise linear activation functions allows for the estimation of exponentially complex functions with the use of a large number of subnetworks specialized in the classification of similar input examples.  During the training process, these subnetworks avoid overfitting with an implicit regularization scheme based on the fact that they must share their parameters with other subnetworks.
Using this framework, we have made an empirical observation that can improve even more the performance of such models.  
We notice that these models assume a balanced initial distribution of data points with respect to the domain of the piecewise linear activation function.  If that assumption is violated, then the piecewise linear activation units can degenerate into purely linear activation units, which can result in a significant reduction of their capacity to learn complex functions.
Furthermore, as the number of model layers increases, this unbalanced initial distribution makes the model ill-conditioned.
Therefore, we propose the introduction of batch normalisation units into deep feedforward neural networks with piecewise linear activations, which drives a more balanced use of these activation units, where each region of the activation function is trained with a relatively large proportion of training samples.  Also, this batch normalisation promotes the pre-conditioning of very deep learning models.
We show that by introducing maxout and batch normalisation units to the network in network model results in a model that produces classification results that are better than or comparable to the current state of the art in CIFAR-10, CIFAR-100, MNIST, and SVHN datasets.

\end{abstract}

\section{Introduction}

The use of piecewise linear activation units in deep learning models~\cite{montufar2014number,nair2010rectified,maas2013rectifier,he2015delving,srivastava2013compete,goodfellow2013maxout}, such as deep convolutional neural network (CNN)~\cite{krizhevsky2012imagenet}, has produced models that are showing state-of-the-art results in several public databases (e.g., CIFAR-10~\cite{krizhevsky2009learning}, CIFAR-100~\cite{krizhevsky2009learning}, MNIST~\cite{lecun1998gradient} and SVHN~\cite{netzer2011reading}).
These piecewise linear activation units have been the subject of study by Montufar et al.~\cite{montufar2014number} and by Srivastava et al.~\cite{srivastava2014understanding}, and the main conclusions achieved in these works are: 1) the use of a multi-layer composition of piecewise linear activation units allows for an exponential division (in terms of the number of network layers) of the input space~\cite{montufar2014number}; 
2) given that the activation units are trained based on a local competition that selects which region of the activation function a training sample will use, "specialized" subnetworks will be formed by the consistency that they respond to similar training samples (i.e., samples lying in one of the regions produced by the exponential division above)~\cite{srivastava2014understanding}
and 3) even though subnetworks are formed and trained with a potentially small number of training samples, these models are not prone to overfitting because these subnetworks share their parameters, resulting in an implicit regularization of the training process~\cite{montufar2014number,srivastava2014understanding}.  

\begin{figure}[t]
\begin{center}
\includegraphics[width=0.8\columnwidth]{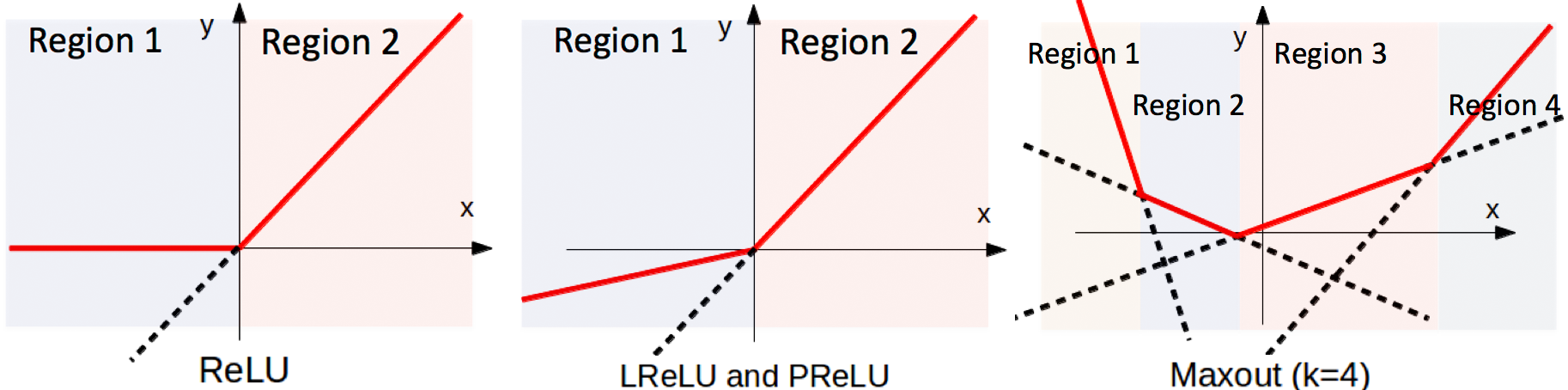} 
\end{center}
\caption{Piecewise linear activation functions: ReLU~\cite{nair2010rectified}, LReLU~\cite{maas2013rectifier}, PReLU~\cite{he2015delving}, and Maxout~\cite{goodfellow2013maxout}.}
\label{fig:intro}
\end{figure}

An assumption made by these works is that a large proportion of the regions of the piecewise linear activation units are active during training and inference.
For instance, in the Rectifier Linear Unit (ReLU)~\cite{nair2010rectified}, Leaky-ReLu (LReLU)~\cite{maas2013rectifier} and Parametric-ReLU (PReLU)~\cite{he2015delving}, there must be two sets of points: one lying in the negative side and another on the positive side of the activation function domain (see region 1 covering the negative side and region 2 on the positive side in $\{\text{P,L}\}$ReLU cases of Fig.~\ref{fig:intro}).  Moreover, in the 
Maxout~\cite{goodfellow2013maxout} and Local winner takes all (LWTA)~\cite{srivastava2013compete} activation units, there must be $k$ sets of points - each set lying in one of the $k$ regions of the activation function domain (see Maxout case in Fig.~\ref{fig:intro}).
This assumption is of utmost importance because if violated, then the activation units may degenerate into simple linear functions that are not capable of exponentially dividing the input space or training the "specialized" subnetworks (i.e., the model capacity is reduced).
Moreover, in learning models that have very deep architectures, the violation of this assumption makes the model ill-conditioned, as shown in the toy example below.
In this paper, we propose the introduction of batch normalisation units~\cite{ioffe2015batch} before the piecewise linear activation units to guarantee 
that the input data is evenly distributed with respect to the activation function domain, which results in a more balanced use of all the regions of the piecewise linear activation units and pre-conditions the model.
Note that Goodfellow et al.~\cite{goodfellow2013maxout} have acknowledged this assumption and proposed the use of dropout~\cite{srivastava2014dropout} to regularize the training process, but dropout cannot guarantee a more balanced distribution of the input data in terms of the activation function domain.  Furthermore, dropout is a regularization technique that does not help pre-condition the model.   Therefore, the issues that we have identified remains with dropout.

\begin{figure*}[t]
\begin{center}
\begin{tabular}{c}
\includegraphics[width=0.7\textwidth]{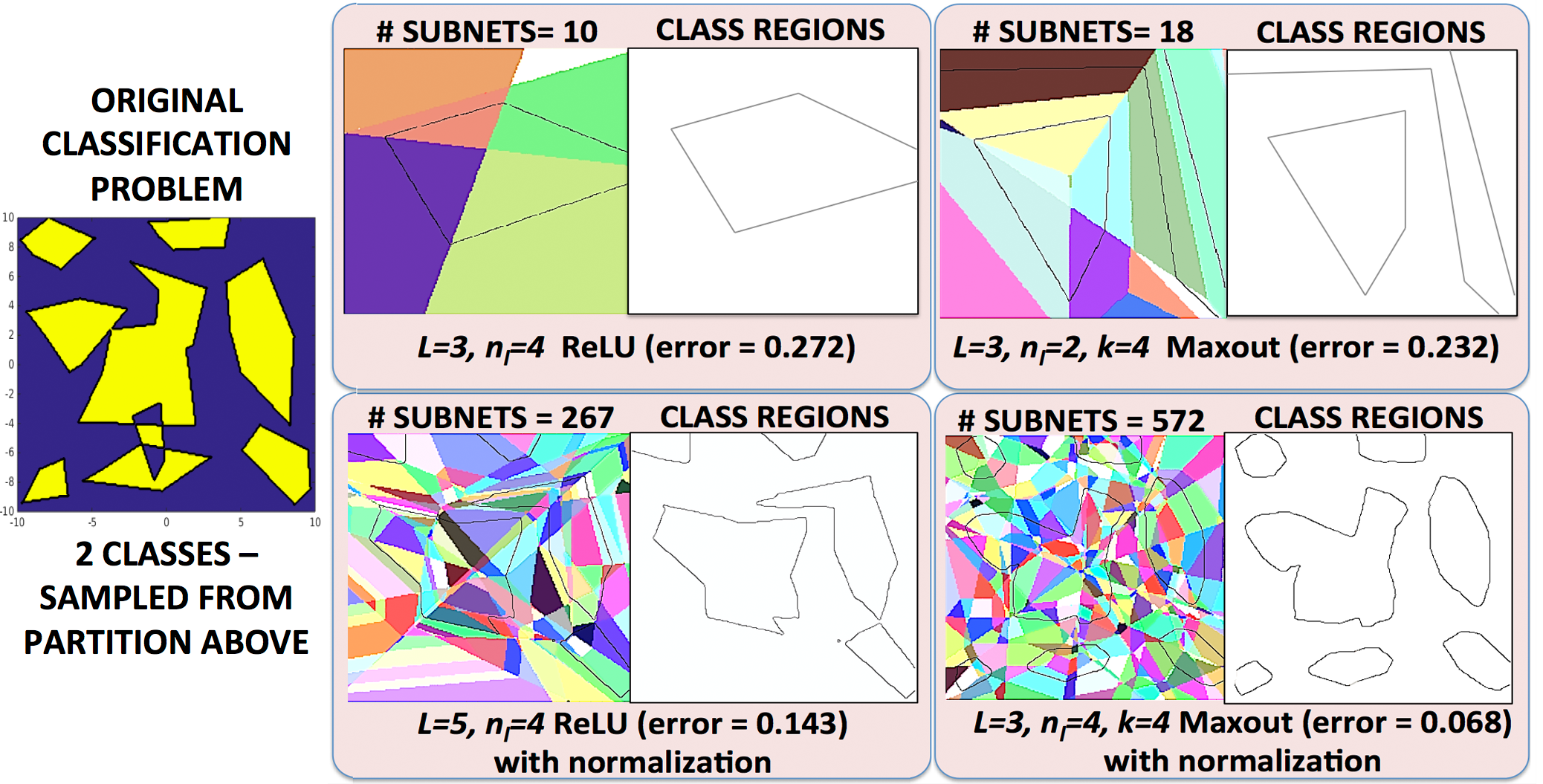} \\
a) Original classification problem (left)  with the liner regions found by each model (represented by the \\
color of each subnet) and classification division of the original space (class regions).\\
\includegraphics[width=0.9\textwidth]{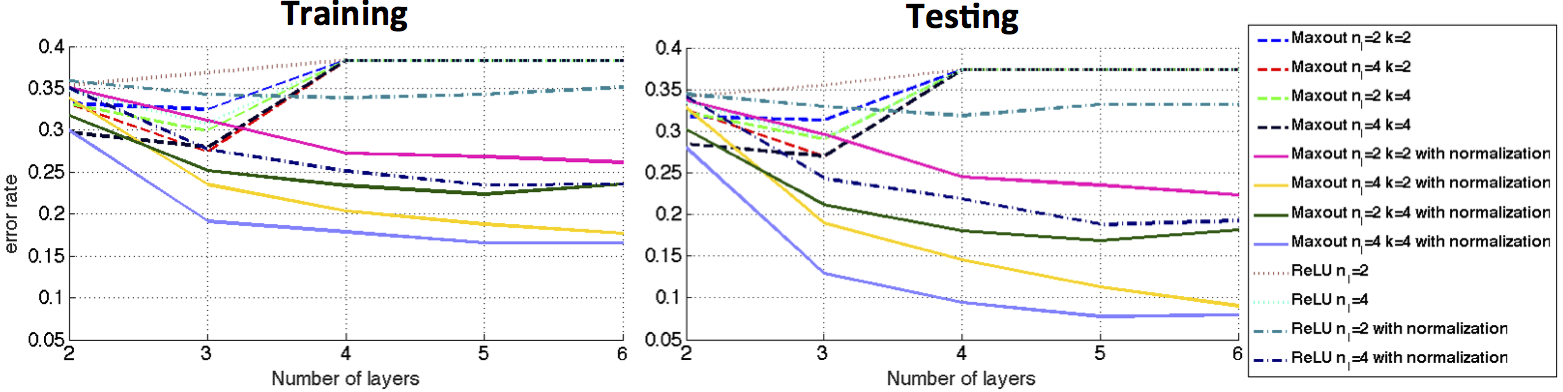} \\
b) Train and test error as a function of the number of layers, number of nodes per layer, piecewise linear \\
activation function, number of regions in the activation function, and the use of normalisation
\end{tabular}
\end{center}
\caption{Toy problem with the division of the space into linear regions and classification profile produced by each model (a), and a quantitative comparison between models (b).}
\label{fig:toy}
\end{figure*}

In order to motivate our observation and the model being proposed in this paper, we show a toy problem that illustrates well our points. Assume that we have a 2-D binary problem, where samples are drawn (12K for training and 2K for testing) using a uniform distribution between $[-10,10]$ (in each dimension) from the partition shown in Fig.~\ref{fig:toy}-(a) (leftmost image), with the colors blue and yellow indicating the class labels.  We train a multi-layer perceptron (MLP) with varying number of nodes per layer $n_l \in \{2,4\}$ and varying number of layers $L \in \{2,3,4,5,6\}$, and it is possible to place two types of piecewise linear activation functions after each layer: ReLU~\cite{nair2010rectified} and maxout~\cite{goodfellow2013maxout}, where for maxout we can vary the number of regions $k \in \{2,4\}$ (e.g., Fig.~\ref{fig:intro} shows a maxout with 4 regions).  Also, before each activation function, we have the option of placing a batch normalisation unit~\cite{ioffe2015batch}.  Training is based on backpropagation~\cite{rumelhart1988learning} using mini-batches of size 100, learning rate of 0.0005 for 20 epochs then 0.0001 for another 20 epochs, momentum of 0.9 and weight decay of 0.0001, where we run five times the training (with different training and test samples) and report the mean train and test errors.  
Finally, the MLP weights are initialized with Normal distribution scaled by 0.01 for all layers.

Analysing the mean train and test error in Fig.~\ref{fig:toy}-(b), we first notice that all models have good generalization capability, which is a characteristic already identified for deep networks that use piecewise linear activation units~\cite{montufar2014number,srivastava2014understanding}.
Looking at the curves for the networks with 2 and 3 layers, where all models seem to be trained properly (i.e., they are pre-conditioned), the models containing batch normalisation units (denoted by "with normalisation") produce the smallest train and test errors, indicating the higher capacity of these models.  Beyond 3 layers, the models that do not use the batch normalisation units become ill-conditioned, producing errors of $0.39$, which effectively means that all points are classified as one of the binary classes. 
In general, batch normalisation allows the use of maxout in deeper MLPs that contain more nodes per layer, and the maxout function contains more regions (i.e., larger $k$).  The best result (in terms of mean test and train error) is achieved with an MLP of 5 or more layers, where each layer contains 4 nodes and maxout has 4 regions (test error saturates at 0.07).  The best results with ReLU are also achieved with batch normalisation, using a large number of layers (5 or more), and 4 nodes per layer, but notice that the smallest ReLU errors (around 0.19 on test set) are significantly higher than the maxout ones, indicating that maxout has larger capacity.
The images in Fig.~\ref{fig:toy}-(a) show the division of the input space (into linear regions) used to train the subnetworks within the MLP model (we show the best performing models of ReLU with and without normalisation and maxout with and without normalisation), where it is worth noticing that the best maxout model (bottom-right image) produces a very large number of linear regions, which generate class regions that are similar to the original classification problem.  The input space division, used to train the subnetworks, are generated by clustering the training points that produce the same activation pattern from all nodes and layers of the MLP.  We also run these same experiments using dropout (of 0.2), and the relative results are similar to the ones presented in Fig.~\ref{fig:toy}-(b), but the test errors with dropout are around $2 \times$ larger, which indicate that dropout does not pre-condition the model (i.e., the models that do not have the batch normalisation units still become ill-conditioned when having 3 or more layers), nor does it balance the input data for the activation units (i.e., the capacity of the model does not increase with dropout).

\begin{figure}[t]
\begin{center}
\includegraphics[width=0.8\columnwidth]{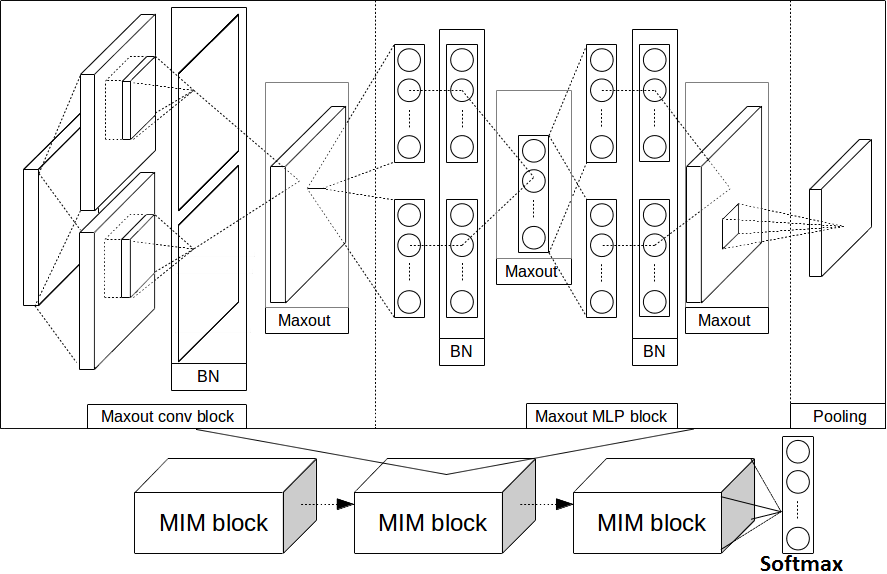}
\end{center}
\caption{Proposed MIM model. The MIM model is based on the NIN~\cite{lin2013network} model. This model contains three blocks that have nearly identical architectures, with small differences in terms of the number of filters and stride in convolution layers. The first two blocks use max pooling and the third block uses average pooling. }
\label{fig:MIM}
\end{figure}

This toy experiment motivates us to propose a model that: 1) contains a large number of layers and nodes per layer, 2) uses maxout activation function~\cite{goodfellow2013maxout}, and 3) uses a batch normalisation unit~\cite{ioffe2015batch} before each maxout layer.  More specifically, we extend the \textit{Network in Network} (NIN) model~\cite{lin2013network}, where we replace the original ReLU units by batch normalisation units followed by maxout units.  Replacing ReLU by maxout has the potential to increase the capacity of the model~\cite{goodfellow2013maxout}, and as mentioned before, the use of batch normalisation units will guarantee a more balanced distribution of this input data for those maxout units, which increases the model capacity and pre-conditions the model.  We call our proposed model the maxout network in maxout network (MIM) - see Fig.~\ref{fig:MIM}.  
We assess the performance of our model on the following datasets:  CIFAR-10~\cite{krizhevsky2009learning}, CIFAR-100~\cite{krizhevsky2009learning}, MNIST~\cite{lecun1998gradient} and SVHN~\cite{netzer2011reading}.  
We first show empirically the improvements achieved with the introduction of maxout and batch normalisation units to the NIN model~\cite{lin2013network}, forming our proposed MIM model, then we show a study on how this model provides a better pre-conditioning for the proposed deep learning model, and finally we show the final classification results on the datasets above, which are compared to the state of the art and demonstrated to be the best in the field in two of these datasets (CIFAR-10, CIFAR-100) and competitive on MNIST and SVHN.

\section{Batch Normalised Deep Learning with Piecewise Linear Activation Units}

In this section, we first explain the piecewise linear activation units, followed by an introduction of how the batch normalisation unit works and a presentation of the proposed MIM model, including its training and inference procedures.

The nomenclature adopted in this section is the same as the one introduced by Montufar et al.~\cite{montufar2014number}, where a feedforward neural network is defined by the function $F:\mathbb R^{n_0} \rightarrow \mathbb R^{\text{out}}$:
\begin{equation}
F(\mathbf{x},\theta) = f_{\text{out}} \circ g_{L} \circ h_L \circ f_L \circ ... \circ g_1 \circ h_1 \circ f_1(\mathbf{x}),
\label{eq:def_CNN}
\end{equation}
where $f(.)$ represents a preactivation function, the parameter $\theta$ is formed by the input weight matrices $\mathbf{W}_{l} \in \mathbb R^{k.n_l \times n_{l-1}}$, bias vectors $\mathbf{b}_l \in \mathbb R^{k.n_l}$ and normalisation parameters $\gamma_l$ and $\beta_l$ in (\ref{eq:normalisation}) for each layer $l \in \{1,...,L\}$, $h_l(.)$ represents a normalisation function, and $g_l(.)$ is a non-linear activation function.  The preactivation function is defined by $f_l(\mathbf{x}_{l-1}) = \mathbf{W}_{l}\mathbf{x}_{l-1}+\mathbf{b}_{l}$, where the output of the $(l-1)^{th}$ layer is $\mathbf{x}_{l} = [ \mathbf{x}_{l,1},...,\mathbf{x}_{l,n_l} ]$, denoting the activations $\mathbf{x}_{l,i}$ of the units $i \in \{1,...,n_l\}$ from layer $l$.  This output is computed from the activations of the preceding layer by $\mathbf{x}_l = g_l(h_l(f_l(\mathbf{x}_{l-1})))$.  Also note that $\mathbf{f}_l = [\mathbf{f}_{l,1},...,\mathbf{f}_{l,n_l}]$ is an array of $n_l$ preactivation vectors $\mathbf{f}_{l,i} \in \mathbb R^{k}$, which after normalisation, results in an array of $n_l$ normalised vectors $\mathbf{h}_{l,i} \in \mathbb R^{k}$ produced by $h_{l,i}(f_{l,i}(\mathbf{x}_{l-1}))$, and the activation of the $i^{th}$ unit in the $l^{th}$ layer is represented by $\mathbf{x}_{l,i} = g_{l,i}(h_{l,1}(f_{l,i}(\mathbf{x}_{l-1})))$.

\subsection{Piecewise Linear Activation Units}
\label{sec:piece_lin}

By dropping the layer index $l$ to facilitate the nomenclature, the recently proposed piecewise linear activation units 
ReLU~\cite{nair2010rectified}, LReLU~\cite{maas2013rectifier}, PReLU~\cite{he2015delving}, and Maxout~\cite{goodfellow2013maxout} are represented as follows~\cite{montufar2014number}:
\begin{equation}
\begin{array}{ll}
\text{ReLU:} & g_i(\mathbf{h}_i) = \max\{ 0, \mathbf{h}_i\}, \\
\text{LReLU or PReLU:} & g_i(\mathbf{h}_i) = \max\{ \alpha . \mathbf{h}_i, \mathbf{h}_i\}, \\
\text{Maxout:} & g_i(\mathbf{h}_i) = \max\{ \mathbf{h}_{i,1},...,\mathbf{h}_{i,k}\}. \\
\end{array}
\label{eq:active_units}
\end{equation}
where $\mathbf{h}_i \in \mathbb R \text{ and } k=1$ for ReLU, LReLU~\cite{maas2013rectifier}, and PReLU~\cite{he2015delving}, $\alpha$ is represented by a small constant in LReLU, but a learnable model parameter in PReLU, $k$ denotes the number of regions of the maxout activation function, and $\mathbf{h}_i=[\mathbf{h}_{i,1},...,\mathbf{h}_{i,k}] \in \mathbb R^k$.  

According to Montufar et al.~\cite{montufar2014number}, the network structure is defined by the input dimensionality $n_0$, the number of layers $L$ and the width $n_l$ of each layer.  A linear region of the function $F:\mathbb R^{n_0} \rightarrow \mathbb R^m$ is a maximal connected subset of $\mathbb R^{n_0}$.  Note from (\ref{eq:active_units}) that rectifier units have two behaviour types: 1) constant 0 (ReLU) or linear (LReLU or PReLU) with a small slope when the input is negative; and 2) linear with slope 1 when input is positive.  These two behaviours are separated by a hyperplane (see Fig.~\ref{fig:intro}) and the set of all hyperplanes within a rectifier layer forms a hyperplane arrangement, which split the input space into several linear regions.  
A multi-layer network that uses rectifier linear units with $n_0$ inputs and $L$ hidden layers with $n \geq n_0$ nodes can compute functions that have $\Omega \left ( (n/n_0)^{L-1} n^{n_0} \right )$ linear regions, and a multi-layer network that uses maxout activation units with $L$ layers of width $n_0$ and rank $k$ can compute functions that have $k^{L-1}k^{n_0}$ linear regions~\cite{montufar2014number}.  These results indicate that multi-layer networks with maxout and rectifier linear units can compute functions with a number of linear regions that grows exponentially with the number of layers~\cite{montufar2014number}.  Note that these linear regions can be observed as the colored polygons in Fig.~\ref{fig:toy}-(a), where the number of linear regions is denoted by "$\#$ SUBNETS".

The training process of networks containing piecewise linear activation units uses a divide and conquer strategy where $\frac{\partial \ell}{\partial \mathbf{W}_l}$ moves the classification boundary for layer $l$ according to the loss function $\ell$ with respect to the points in its current linear region (similarly for the bias term $\mathbf{b}_l$), and $\frac{\partial \ell}{\partial \mathbf{x}_{l-1}}$ moves the offending points (i.e., points being erroneously classified) away from their current linear regions.
Dividing the data points into an exponentially large number of linear regions is advantageous because the training algorithm can focus on minimizing the loss for each one of these regions almost independently of others - this is why we say it uses a divide and conquer algorithm.
We also say that it is an almost independent training of each linear region because the training parameters for each region are shared with all other regions, and this helps the regularization of the training process.
However, the initialization of this training process is critical because if the data points are not evenly distributed at the beginning, then all these points may lie in only one of the regions of the piecewise linear unit.
This will drive the learning of the classification boundary for that specific linear region, where the loss will be minimized for all those points in that region, and the boundary for the other linear regions will be trained less effectively with much fewer points.
This means that even the points with relatively high loss will remain in that initial region because the other regions have  been ineffectively trained, and consequently may have a larger loss. This issue is very clear with the use of maxout units, where in the extreme case, only one of the $k$ regions is active, which means that the maxout unit will behave as a simple linear unit.  If a large amount of maxout units behave as linear units, then this will reduce the ability of these networks to compute functions that have an exponential number of linear regions, and consequently decrease the capacity of the model.

\subsection{Batch Normalisation Units}

In order to force the initialization to distribute the data points evenly in the domain of the piecewise activation functions, such that a large proportion of the $k$ regions is used, we propose the use of batch normalisation by Ioffe and Szegedy's~\cite{ioffe2015batch}.
This normalisation has been proposed because of the difficulty in initializing the network parameters and setting the value for the learning rate, and also because the inputs for each layer are affected by the parameters of the previous layers.  These issues lead to a complicated learning problem, where the input distribution for each layer changes continuously - an issue that is called covariate shift~\cite{ioffe2015batch}.
The main contribution of this batch normalisation is the introduction of a simple \emph{feature-wise} centering and normalisation to make it have mean zero and variance one, which is followed by a batch normalisation (BN) unit that shifts and scales the normalised value.  For instance, assuming that the input to the normalisation unit is $\mathbf{f} = [\mathbf{f}_1,...,\mathbf{f}_{n_l}]$, where $\mathbf{f}_i \in \mathbb R^{k}$, the BN unit consists of two stages:
\begin{equation}
\begin{array}{ll}
\text{Normalisation: }  &  \hat{\mathbf{f}}_{i,k} = \frac{\mathbf{f}_{i,k} - E[\mathbf{f}_{i,k}]}{\sqrt{\text{Var}[\mathbf{f}_{i,k}]}} \\
\text{Scale and shift: }  &  \mathbf{h}_{i,k} =   \gamma_i \hat{\mathbf{f}}_{i,k} + \beta_i  \\
\end{array},
\label{eq:normalisation}
\end{equation}
where the shift and scale parameters $\{ \gamma_i,\beta_i \}$ are new network parameters that participate in the training procedure~\cite{ioffe2015batch}.  Another important point is that the BN unit does not process each training sample independently, but it uses both the training sample and other samples in a mini-batch.

\subsection{Maxout Network in Maxout Network Model}

As mentioned in Sec.~\ref{sec:piece_lin}, the number of linear regions that networks with piecewise linear activation unit can have grows exponentially with the number of layers, so it is important to add as many layers as possible in order to increase the ability of the network to estimate complex functions.
For this reason, we extend the recently proposed Network in Network (NIN)~\cite{lin2013network} model, which is based on a CNN that uses a multi-layer perceptron (MLP) as its activation layer (this layer is called the Mlpconv layer).  In its original formulation, the NIN model introduces the Mlpconv with ReLU activation after each convolution layer, and replaces the fully connected layers for classification in CNN (usually present at the end of the whole network) by a spatial average of the feature maps from the last Mlpconv layer, which is fed into a softmax layer.  
In particular, we extend the NIN model by replacing the ReLU activation after each convolution layer of the Mlpconv by a maxout activation unit, which has the potential to increase even further the model capacity.  In addition, we also add the BN unit before the maxout units.  These two contributions form our proposed model, we give it a simple name Maxout Network in Maxout Network Model (MIM), which is depicted in Fig.~\ref{fig:MIM}.
Finally, we include a dropout layer~\cite{srivastava2014dropout} between MIM blocks for regularizing the model.

\begin{table*}[t]
\begin{center}
\begin{tabularx}{\textwidth}{|c|c|c|c|c|c|c|}
\hline
Arch. 	&  m-conv1 & m-mlp1 & m-conv2 & m-mlp2 & m-conv3 & m-mlp3 \\
\hline
\hline
\parbox[c]{0.105\textwidth}{ \tiny\shortstack{CIFAR-10\\CIFAR-100\\SVHN} } & 
\parbox[c]{0.12\textwidth}{ \tiny\shortstack{5x5x192 \\ stride. 1, pad. 2, k. 2 \\ BN}} &
\parbox[c]{0.12\textwidth}{ \tiny\shortstack{1x1x160 \\ stride. 1, pad. 0, k. 2 \\ BN \\ $\downarrow$ \\ 1x1x96 \\ stride. 1, pad. 0, k. 2 \\ BN \\ 3x-max.pool \\ dropout  }} & 
\parbox[c]{0.12\textwidth}{ \tiny\shortstack{5x5x192 \\ stride. 1, pad. 2, k. 2 \\ BN}} &
\parbox[c]{0.12\textwidth}{ \tiny\shortstack{1x1x192 \\ stride. 1, pad. 0, k. 2 \\ BN \\ $\downarrow$ \\ 1x1x192 \\ stride. 1, pad. 0, k. 2 \\ BN \\ 3x-max.pool \\ dropout  }} &
\parbox[c]{0.12\textwidth}{ \tiny\shortstack{3x3x192 \\ stride. 1, pad. 0, k. 2 \\ BN}} &
\parbox[c]{0.12\textwidth}{ \tiny\shortstack{1x1x160 \\ stride. 1, pad. 0, k. 2 \\ BN \\ $\downarrow$ \\ 1x1x10(100) \\ stride. 1, pad. 0, k. 2 \\ BN \\ 8x-avg.pool  }} 
\\
\hline
\hline
\parbox[c]{0.09\textwidth}{ \tiny{MNIST} } & 
\parbox[c]{0.115\textwidth}{ \tiny\shortstack{5x5x128 \\ stride. 1, pad. 2, k. 2 \\ BN}} &
\parbox[c]{0.115\textwidth}{ \tiny\shortstack{1x1x96 \\ stride. 1, pad. 0, k. 2 \\ BN \\ $\downarrow$ \\ 1x1x48 \\ stride. 1, pad. 0, k. 2 \\ BN \\ 3x-max.pool \\ dropout  }} & 
\parbox[c]{0.115\textwidth}{ \tiny\shortstack{5x5x128 \\ stride. 1, pad. 2, k. 2 \\ BN}} &
\parbox[c]{0.115\textwidth}{ \tiny\shortstack{1x1x96 \\ stride. 1, pad. 0, k. 2 \\ BN \\ $\downarrow$ \\ 1x1x48 \\ stride. 1, pad. 0, k. 2 \\ B \\ 3x-max.pool \\ dropout  }} &
\parbox[c]{0.115\textwidth}{ \tiny\shortstack{3x3x128 \\ stride. 1, pad. 0, k. 2 \\ BN}} &
\parbox[c]{0.115\textwidth}{ \tiny\shortstack{1x1x96 \\ stride. 1, pad. 0, k. 2 \\ BN \\ $\downarrow$ \\ 1x1x10 \\ stride. 1, pad. 0, k. 2 \\ BN \\ 7x-avg.pool  }} 
\\
\hline
\end{tabularx}
\end{center}
\caption{The proposed MIM model architectures used in the experiments. In each maxout-conv unit (m-conv), the convolution kernel is defined by the first row in the block: (height)x(width)x(num of units). The second row of the block contains the information of convolution stride (stride), padding (pad), and maxout rank ($k$). The third row contains the BN units.  Each layer of the maxout-mlp unit (m-mlp) is equivalent to a maxout-conv unit with 1x1 convolution kernel size. A softmax layer is present as the last layer of the model (but not shown in this table).  The model on top row is used on CIFAR-10 and 100 and SVHN, while the model on the bottom row is for MNIST.}
\label{tab:mim_spec}
\end{table*}

\section{Experiments}

We evaluate our proposed method on four common deep learning benchmarks: CIFAR-10~\cite{krizhevsky2009learning}, CIFAR-100~\cite{krizhevsky2009learning}, MNIST~\cite{lecun1998gradient} and SVHN~\cite{netzer2011reading}.
The CIFAR-10~\cite{krizhevsky2009learning} dataset contains 60000 32x32 RGB images of 10 classes of common visual objects (e.g., animals, vehicles, etc.), where 50000 images are for training and the rest 10000 for testing. 
The CIFAR-100~\cite{krizhevsky2009learning} is an extension of CIFAR-10, where the difference is that CIFAR-100 has 100 classes with 500 training images and 100 testing images for each class. 
In both CIFAR-10 and 100, the visual objects  are well-centred in the images.
The MNIST~\cite{lecun1998gradient} dataset is a standard benchmark for comparing learning methods. It contains 70000 28x28 grayscale images of numerical digits from 0 to 9, divided as 60000 images for training and 10000 images for testing. 
Finally, the Street View House Number (SVHN)~\cite{netzer2011reading} dataset is a real-word digit dataset with over 600000 32x32 RGB images containing images of house numbers (i.e., digits 0-9). The cropped digits are well-centred and the original aspect ratio is kept, but some distracting digits are present next to the centred digits of interest. The dataset is partitioned into training, test and extra sets, where the extra 530000 images are less difficult samples to be used as extra training data.

For each of these datasets, we validate our algorithm using the same training and validation splitting described by Goodfellow et al.~\cite{goodfellow2013maxout} in order to estimate the model hyper-parameters. 
For the reported results, we run 5 training processes, each with different model initializations, and the test results consist of the mean and standard deviation of the errors in these 5 runs. 
Model initialization is based on randomly producing the MIM weight values using a Normal distribution, which is  multiplied by $0.01$ in the first layer of the first MIM block and by $0.05$ in all remaining layers.
Moreover, we do not perform data augmentation for any of these datasets and only compare our MIM model with the state-of-the-art methods that report non data-augmented results. 
For the implementation, we use the MatConvNet~\cite{vedaldi2014matconvnet} CNN toolbox and run our experiments on a standard PC equipped with Intel i7-4770 CPU and Nvidia GTX TITAN X GPU. 
Finally, Tab.~\ref{tab:mim_spec} specifies the details of the proposed MIM architecture used for each dataset.

Below, we first show experiments that demonstrate the performance of the original NIN model~\cite{lin2013network} with the introduction of maxout and BN units, which comprise our contributions in this paper that form the MIM model.  Then, we show s study on how the BN units pre-conditions the NIN model.  Finally, we show a comparison between our proposed MIM model against the current state of the art on the aforementioned datasets.

\subsection{Introducing Maxout and Batch Normalisation to NIN model}
\label{sec:mim}

\begin{table}[]
\begin{center}
\resizebox{0.9\columnwidth}{!}{
\begin{tabular}[]{lr}
\hline
Method &  Test Error (mean $\pm$ standard deviation)\\
\hline
NIN~\cite{lin2013network}   &  $10.41\%$ \\
NIN with maxout (without BN) &  $10.95\pm 0.21 \%$ \\
NIN with ReLU (with BN) & $9.43  \pm 0.21 \%$ \\
MIM (= NIN with maxout and BN - our proposal) & $8.52 \pm 0.20 \%$ \\
\hline
\end{tabular}
}
\end{center}
\caption{Results on CIFAR-10 of  the introduction of maxout and BN units to NIN, which produce our proposed MIM model (last row).}
\label{tab:cifar10_exp_component}

\end{table}

In this section, we use CIFAR-10 to show the contribution provided by each component proposed in this paper.  
The first row of Tab.~\ref{tab:cifar10_exp_component} shows the published results of NIN~\cite{lin2013network}. In our first experiment, we replace all ReLU units from the NIN model by the maxout units (with $k=2$) and run the training and test experiments described above (test results are shown in the second row of Tab.~\ref{tab:cifar10_exp_component}).  Second, we include the BN units before each ReLU unit in the NIN model and show the test results in the third row of Tab.~\ref{tab:cifar10_exp_component}.  Finally, we include the maxout and BN units to the NIN model, which effectively forms our proposed MIM model, and test results are displayed in the fourth row of Tab.~\ref{tab:cifar10_exp_component}.  

\subsection{Ill-conditioning Study in Real Datasets}
\label{sec:ill}

\begin{figure}[t]
\begin{center}
\includegraphics[width=\columnwidth]{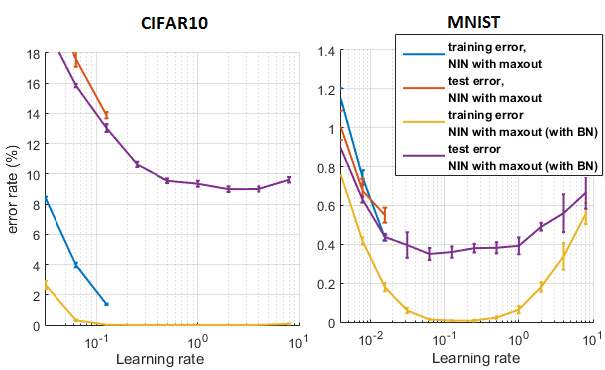}
\end{center}
\caption{The ill-conditioning, measured by the model's inability to converge as a function of the learning rate. The training error (blue curve) and test error (orange curve) of the models trained without BN stay at the initial error when the  learning rate is above a certain value, showing no sign of convergence (the results in terms of these learning rates are therefore omitted).}
\label{fig:ill-cond}
\end{figure}

The study of how the BN units pre-conditions the proposed model (on CIFAR-10 and MNIST) is shown in Fig.~\ref{fig:ill-cond}.
For this evaluation, we use the combination of NIN and maxout units as the standard model, and ensure that the learning rate is the only varying parameter. 
We train five distinct models with learning rates in $[10^{-2}, 10^1]$ for CIFAR-10, and $[10^{-3}, 10^1]$ for MNIST, and plot the error curves with the mean and standard deviation values.
From  Fig.~\ref{fig:ill-cond}, we can see that without BN, a deep learning model can become ill-conditioned. It is also interesting to see that these un-normalized models give best performance right before the learning rate drives it into the ill-conditioning mode.

\subsection{Comparison with the State of the Art}
\label{sec:comparison}

We compare the proposed MIM model (first row of Tab.~\ref{tab:mim_spec}) with Stochastic Pooling~\cite{zeiler2013stochastic}, Maxout Networks~\cite{goodfellow2013maxout}, Network in Network~\cite{lin2013network}, Deeply-supervised nets~\cite{lee2014deeply}, and recurrent CNN~\cite{liang2015recurrent} on CIFAR-10~\cite{krizhevsky2009learning} and show the results in Tab.~\ref{tab:cifar10_exp}.
The comparison on CIFAR-100~\cite{krizhevsky2009learning} against the same state-of-the-art models above, and also the Tree based Priors~\cite{srivastava2013discriminative}, is shown in Tab.~\ref{tab:cifar100_exp}.
The performance on MNIST~\cite{lecun1998gradient} of the MIM model (second row of Tab.~\ref{tab:mim_spec}) is compared against Stochastic Pooling~\cite{zeiler2013stochastic}, Conv. Maxout+Dropout~\cite{goodfellow2013maxout}, Network in Network~\cite{lin2013network}, Deeply-supervised nets~\cite{lee2014deeply}, and recurrent CNN~\cite{liang2015recurrent} in Tab.~\ref{tab:mnist_exp}.
It is important to mention that the best result we observed with the MIM model on MNIST over the 5 runs is $0.32\%$.
Finally, our MIM model in the first row of Tab.~\ref{tab:mim_spec} is compared against the same models above, plus Dropconnect~\cite{wan2013regularization} on SVHN~\cite{netzer2011reading}, and results are displayed in Tab.~\ref{tab:svhn_exp}.

\begin{table}[]
\begin{center}
\resizebox{0.9\columnwidth}{!}{
\begin{tabular}[]{lr}
\hline
Method &  Test Error (mean $\pm$ standard deviation)\\
\hline
Stochastic Pooling~\cite{zeiler2013stochastic} &  $15.13\%$ \\
Maxout Networks~\cite{goodfellow2013maxout} & $11.68\%$ \\
Network in Network~\cite{lin2013network} & $10.41\%$ \\
Deeply-supervised nets~\cite{lee2014deeply} & $9.69\%$ \\
RCNN-160~\cite{liang2015recurrent} & $8.69\%$ \\
MIM (our proposal) & $8.52 \pm 0.20 \%$ \\
\hline
\end{tabular}
}
\end{center}
\caption{Comparison between MIM and the state-of-the-art methods on CIFAR-10~\cite{krizhevsky2009learning}. }
\label{tab:cifar10_exp}
\end{table}

\begin{table}[]
\begin{center}
\resizebox{0.9\columnwidth}{!}{
\begin{tabular}[]{lr}
\hline
Method &  Test Error (mean $\pm$ standard deviation)\\
\hline
Stochastic Pooling~\cite{zeiler2013stochastic} &  $42.51\%$ \\
Maxout Networks~\cite{goodfellow2013maxout} & $38.57\%$ \\
Tree based Priors~\cite{srivastava2013discriminative} & $36.85\%$ \\
Network in Network~\cite{lin2013network} & $35.68\%$ \\
Deeply-supervised nets~\cite{lee2014deeply} & $34.57\%$ \\
RCNN-160~\cite{liang2015recurrent} & $31.75\%$ \\
MIM (our proposal) & $29.20 \pm 0.20 \%$ \\
\hline
\end{tabular}
}
\end{center}
\caption{Comparison between MIM and the state-of-the-art methods on CIFAR-100~\cite{krizhevsky2009learning}. }
\label{tab:cifar100_exp}
\end{table}

\begin{table}[]
\begin{center}
\resizebox{0.9\columnwidth}{!}{
\begin{tabular}[]{lr}
\hline
Method &  Test Error (mean $\pm$ standard deviation)\\
\hline
Stochastic Pooling~\cite{zeiler2013stochastic} & $0.47\%$ \\
Conv. Maxout+Dropout~\cite{goodfellow2013maxout} & $0.47\%$ \\
Network in Network~\cite{lin2013network} & $0.45\%$ \\
Deeply-supervised nets~\cite{lee2014deeply} & $0.39\%$ \\
MIM (our proposal) & $0.35 \pm 0.03\%$ \\ 
RCNN-96~\cite{liang2015recurrent} & $0.31\%$ \\
\hline
\end{tabular}
}
\end{center}
\caption{Comparison between MIM and the state-of-the-art methods on MNIST~\cite{lecun1998gradient}.}
\label{tab:mnist_exp}
\end{table}

\begin{table}[]
\begin{center}
\resizebox{0.9\columnwidth}{!}{
\begin{tabular}[]{lr}
\hline
Method &  Test Error (mean $\pm$ standard deviation)\\
\hline
Stochastic Pooling~\cite{zeiler2013stochastic} & $2.80\%$ \\
Conv. Maxout+Dropout~\cite{goodfellow2013maxout} & $2.47\%$ \\
Network in Network~\cite{lin2013network} & $2.35\%$ \\
MIM (our proposal) & $1.97 \pm 0.08\%$ \\ 
Dropconnect~\cite{wan2013regularization} & $1.94\%$  \\
Deeply-supervised nets~\cite{lee2014deeply} & $1.92\%$ \\
RCNN-192~\cite{liang2015recurrent} & $1.77\%$ \\
\hline
\end{tabular}
}
\end{center}
\caption{Comparison between MIM and the state-of-the-art methods on SVHN~\cite{netzer2011reading}.}
\label{tab:svhn_exp}
\end{table}

\vspace{-.01in}
\section{Discussion and Conclusion}


The results in Sec.~\ref{sec:mim} show that the replacement of ReLU by maxout increases the test error on CIFAR-10, similarly to what has been shown in Fig.~\ref{fig:toy}.  The introduction of BN with ReLU activation units provide a significant improvement of the test error, and the introduction of BN units before the maxout units produce the smallest error, which happens due to the even input data distribution with respect to the activation function domain, resulting in a more balanced use of all the regions of the maxout units.
The study in Sec.~\ref{sec:ill} clearly shows that the introduction of BN units pre-conditions the model, allowing it to use large learning rates and produce more accurate classification.
The comparison against the current state of the art (Sec.~\ref{sec:comparison}) shows that the proposed MIM model produces the best result in the field on CIFAR-10 and CIFAR-100.  On MNIST, our best result over five runs is comparable to the best result in the field.  Finally, on SVHN, our result is slightly worse than the current best result in the field.
An interesting point one can make is with respect to the number of regions $k$ that we set for the MIM maxout units.  Note that we set $k=2$ because we did not notice any significant improvement with bigger values of $k$, and also because the computational time and memory requirements of the training become intractable.

This paper provides an empirical demonstration that the combination of piecewise linear activation units with BN units provides a powerful framework to be explored in the design of deep learning models.  More specifically, our work shows how to guarantee the assumption made in the use of piecewise linear activation units about the balanced distribution of the input data for these units.  This empirical evidence can be shown more theoretically in a future work, following the results produced by Montufar, Srivastava and others~\cite{montufar2014number,srivastava2014understanding}.

{\small
\bibliographystyle{IEEEbib}
\bibliography{egbib}
}
\end{document}